\pdfoutput=1

\documentclass[11pt]{article}

\usepackage[review]{acl}
\usepackage{tabularx}
\usepackage{times}
\usepackage{latexsym}
\usepackage{amsmath}
\usepackage{graphicx}
\usepackage[T1]{fontenc}
\usepackage{multirow}

\usepackage[utf8]{inputenc}
\usepackage{makecell}

\usepackage{microtype}
\usepackage{subfigure}
\usepackage{listings}

\title{HRLAIF: Improvements in Helpfulness and Harmlessness in Open-domain Reinforcement Learning From AI Feedback}

\author{
Ang Li$^{*}$, Qiugen Xiao\thanks{~~Equal contribution.}, \\ \textbf{Peng Cao}, 
 \textbf{Jian Tang}, \textbf{Yi Yuan},  \textbf{Zijie Zhao}, \textbf{Xiaoyuan Chen}, \textbf{Liang Zhang}, \\ \textbf{Yukang Gan},
\textbf{Xiangyang Li},   \textbf{Kaitong Yang},  \textbf{Weidong Guo}, \textbf{Yu Xu}, \textbf{Daniell Wang}, \textbf{Ying Shan} \\
 Tencent PCG
}

\begin{document}
\maketitle
\begin{abstract}
Reinforcement Learning from AI Feedback (RLAIF) has the advantages of shorter annotation cycles and lower costs over Reinforcement Learning from Human Feedback (RLHF), making it highly efficient during the rapid strategy iteration periods of large language model (LLM) training. Using ChatGPT as a labeler to provide feedback on open-domain prompts in RLAIF training, we observe an increase in human evaluators' preference win ratio for model responses, but a decrease in evaluators' satisfaction rate. Analysis suggests that the decrease in satisfaction rate is mainly due to some responses becoming less helpful, particularly in terms of correctness and truthfulness, highlighting practical limitations of basic RLAIF. In this paper, we propose Hybrid Reinforcement Learning from AI Feedback (HRLAIF). This method enhances the accuracy of AI annotations for responses, making the model's  helpfulness more robust in training process. Additionally, it employs AI for Red Teaming, further improving the model's harmlessness. Human evaluation results show that HRLAIF inherits the ability of RLAIF to enhance human preference for outcomes at a low cost while also improving the satisfaction rate of responses. Compared to the policy model before Reinforcement Learning (RL), it achieves an increase of 2.08\% in satisfaction rate, effectively addressing the issue of a decrease of 4.58\% in satisfaction rate after basic RLAIF.

\end{abstract}
\section{Introduction}
Reinforcement Learning from Human Feedback (RLHF) \citep{ouyang2022training, bai2022training} has been proven effective by recent studies in aligning the responses of large language models (LLMs) \citep{vaswani2017attention, bender2021dangers} with human preferences. However, the reward model (RM) used to train the policy model in RLHF relies on human preference labeling for language model responses, which is a costly and time-consuming process. To address this issue, some researchers propose using AI to provide feedback for AI, namely the RLAIF method \citep{bai2022constitutional}. Compared to RLHF, RLAIF has the advantages of lower cost and shorter in cycles.

We use ChatGPT\cite{Liu_2023} as a labeler for aligning language models and find that after RLAIF, the model's responses have a higher win ratio in human preference comparison. This indicates that RLAIF indeed has the advantage of enhancing human preferences at a lower cost. However, we also identified a decrease in the human satisfaction rate of responses with this method.

\begin{figure}
	\centering
	\subfigure[\label{fig:a} Satisfaction rate]{
		\includegraphics[scale=0.16]{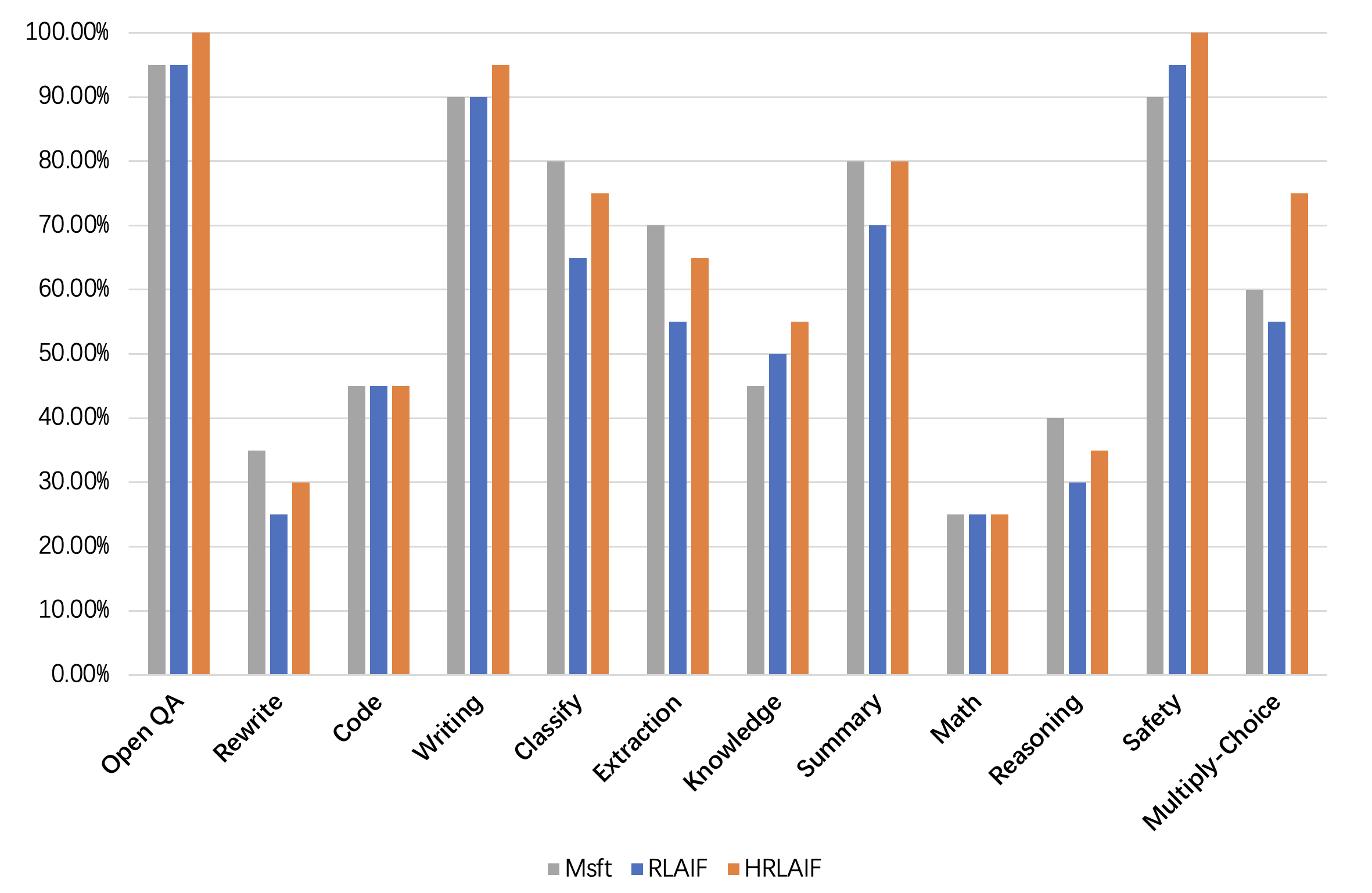}}
        \subfigure[\label{fig:b} Preference win ratio]{
		\includegraphics[scale=0.16]{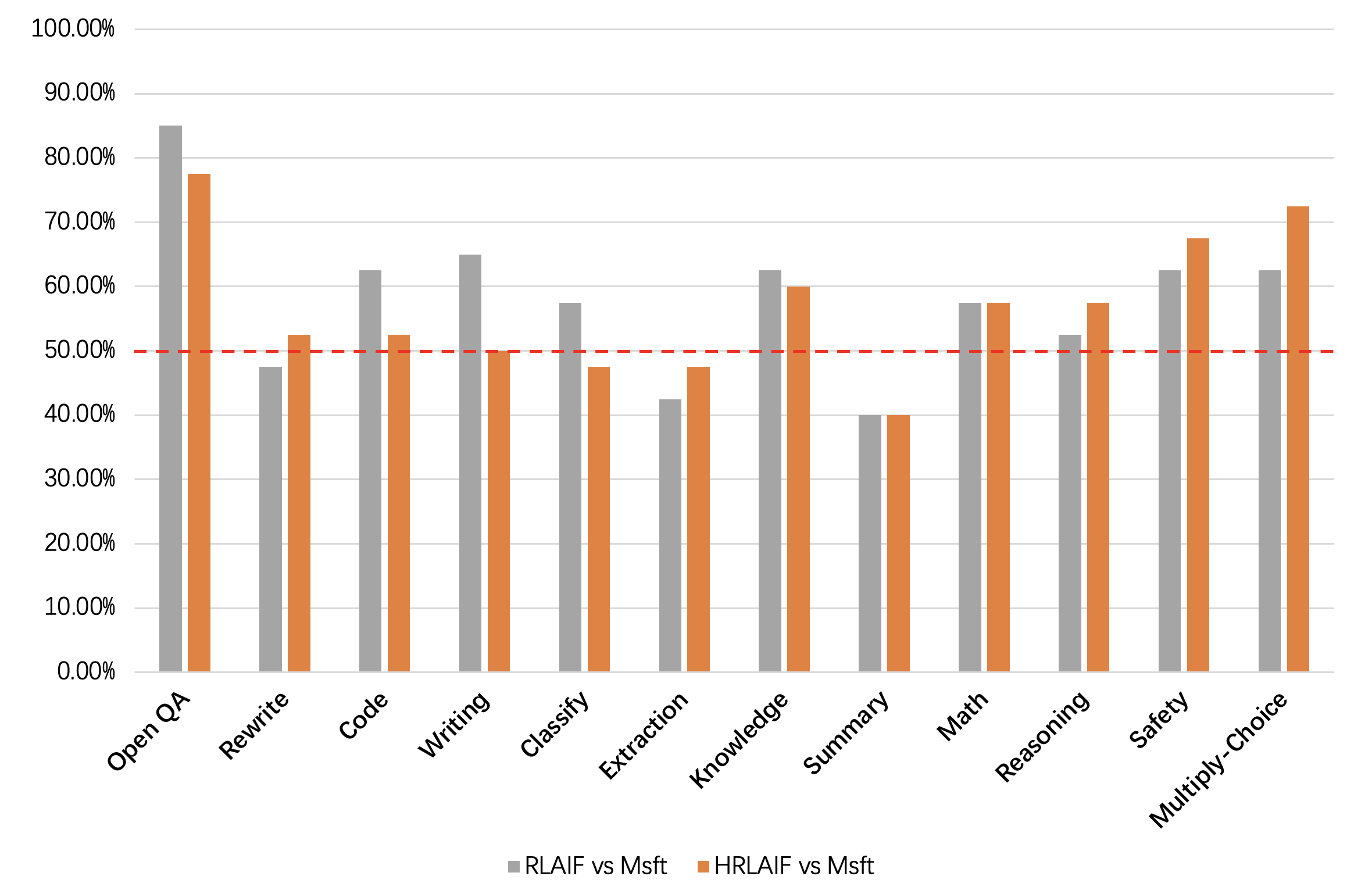}}
	\\
	\caption{Human evaluation results of basic RLAIF and HRLAIF on different prompt categories.}
	\label{fig: human} 
\end{figure}

\begin{figure*} [ht]
    \centering
    \includegraphics[width=\linewidth]{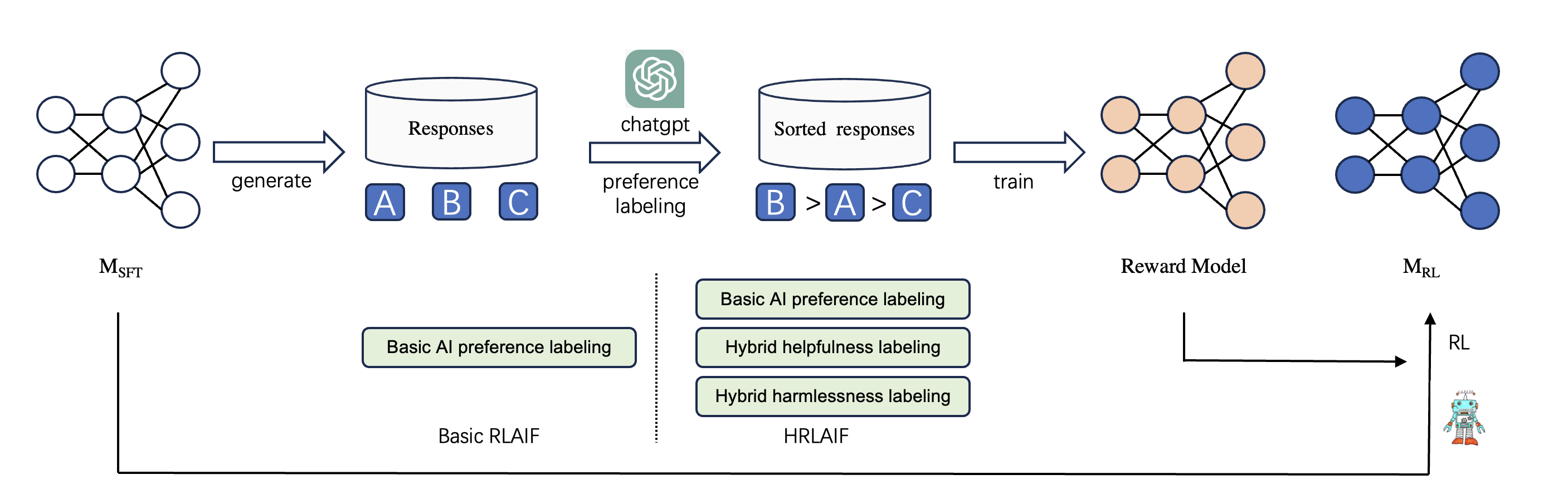}
    \caption{Framework of basic RLAIF and HRLAIF.}
    \label{fig:framework}
\end{figure*}

Upon analysis, the increase in preference is primarily attributed to improvements in the stylistic paradigms of model responses. The decrease in satisfaction rate, on the other hand, is due to some responses becoming less helpful, particularly in terms of correctness and truthfulness. This issue arises mainly because AI has a lower accuracy in preference annotation for certain types of tasks, leading to the RM trained with AI feedback being ineffective in judging the correctness of responses.

To address the issues identified in RLAIF, we propose a novel method called Hybrid Reinforcement Learning from AI Feedback (HRLAIF). The term 'Hybrid' in this context refers to the phased annotating on different prompt categories during the AI preference labeling process. This method significantly enhances the reliability of AI annotations on certain categories of prompts, leading to a more robust model in helpfulness during Reinforcement Learning (RL).

Moreover, HRLAIF also includes using AI for Red Teaming with a toxic prompt set. This approach further enhances the harmlessness of the model.

In this paper, we primarily implement hybrid AI preference labeling in prompt categories such as math computation, multiple-choice question, and toxic prompt rejection, followed by RL training. Human evaluation results on prompts of 12 categories show that both basic RLAIF and HRLAIF can enhance the model's win ratio in human preference comparison after training. However, in terms of the response satisfaction rate, compared to the policy model before training, basic RLAIF experiences a decrease of 4.58\%, while HRLAIF achieves an increase of 2.08\%. Our main contributions are as follows:

\begin{itemize}
    \item We propose a novel method, HRLAIF, which addresses the issue of decreased helpfulness observed in the basic RLAIF process, while also further enhancing model's harmlessness.
    \item We quantify the effectiveness of the above approaches with popular LLM benchmarks and human evaluations based on our Chinese multi-category evaluation set. In benchmarks for helpfulness and harmlessness, HRLAIF outperforms basic RLAIF. In human evaluations, both basic RLAIF and HRLAIF show an increase in the win ratio of human preference comparison after training. However, while basic RLAIF experiences a  decrease of 4.58\% in response satisfaction rate, HRLAIF achieves an increase of 2.08\%.
\end{itemize}

\section{Related Works}
\label{sec:Related Work}
\noindent \textbf{LLM learning from human feedback.}  \citet{christiano2017deep} explored goals defined in terms of human preferences between pairs of trajectory segments. \citet{9054379} used on-policy learning and model the user emotional state to improve a seq2seq model. \citet{nahian2021training} introduced an approach to value aligned reinforcement learning, and trained an agent with human reward signals. \citet{ouyang2022training} showed an avenue for aligning LLMs with user intent on a wide range of tasks by fine-tuning with human feedback. \citet{touvron2023llama} developed Llama 2, which was trained with batches of human preference data annotation in RLHF fine-tuning stage. There were also studies that optimize the difference in log probabilities between winning and losing responses based on human feedback results \cite{rafailov2023direct, yuan2023rrhf}.

\noindent \textbf{LLM Reinforcement Learning.}  Best-of-N(or reject sampling) uses the reward model to select the highest-scoring responses out of n outputs generated by the policy model, and then fine-tunes the policy model with these responses \cite{stiennon2020learning,askell2021general,touvron2023llama}. Proximal Policy Optimization (PPO) maximizes model response rewards with advantage function, and uses KL penalty to avoid over-optimization \cite{schulman2017proximal, ouyang2022training}. \citet{lu2022quark} introduced Quark, an algorithm for optimizing a reward function that quantifies an wanted property.

\noindent \textbf{RLAIF.} Training AI with RL from AI Feedback was firstly proposed by \citet{bai2022constitutional}. They trained a harmless AI assistant through self-improvement, without any human labels identifying harmful outputs. \citet{dubois2024alpacafarm} designed LLM prompts to simulate human feedback using chat-GPT and GPT-4, and contributed reference implementations for several RL methods. \citet{lee2023rlaif} conducted studies on techniques for generating aligned AI preferences for summarization tasks and achieved human-level performance. \citet{starling2023}  instructed GPT-4 to  conduct pairwise comparisons for all response pairs to reduced the positional bias of GPT. 

In this paper, we implements a RLAIF approach using GPT for AI preference labeling on Chinese data. Additionally, we find some shortcomings when GPT is directly used as labeler and propose corresponding solutions.In terms of RL algorithms, we choose the popular PPO algorithm for RL in this paper, as it seems to have a higher performance ceiling compared to Best-of-N. Theoretically, other RL algorithms could also be applicable to HRLAIF proposed in this paper.

\section{Methodology}

This chapter provides a detailed description of basic RLAIF and HRLAIF. The framework of basic RLAIF and HRLAIF is plotted in Figure \ref{fig:framework}. Their complete process includes three stages: AI preference labeling, RM training, and RL. Compared to basic RLAIF, HRLAIF primarily enhances the reliability of AI labeling through the implementation of hybrid AI preference labeling, which leads to improvements in the model's helpfulness and harmlessness after RL.

In this paper, we use $M_{SFT}$ to denote the policy model that has undergone Supervised Fine-Tuning (SFT) and requires RL. $L_{AI}$ denotes the AI assistant we use for preference labeling.

\subsection{Basic AI Preference Labeling}

Basic AI preference labeling refers to the direct use of AI for preference annotation. Following existing works in Section \ref{sec:Related Work}, our basic AI preference labeling process outcomes preference partial order of two different responses upon one prompt \cite{zhou2022large}. To enhance the ability of preference labeling, we primarily incorporate strategies such as Position Bias Eliminating and Chain of Thought.

\noindent \textbf{Position Bias Eliminating.} Existing studies have confirmed the presence of Position Bias in the LLMs' preference evaluation, meaning that swapping the order of two responses in the context can influence preference outcomes provided by $L_{AI}$ \cite{Wang2023LargeLM,starling2023,lee2023rlaif}. Therefore, for each pair to be compared, we perform two comparisons with swapped response orders, and then decide their final partial order based on the average score of each response across both comparisons.

\noindent \textbf{Chain of Thought.} This method has been designed to enhance the model's ability to solve complex problems that require multi-step reasoning\cite{wei2023chainofthought}. We require $L_{AI}$ to thoroughly think and compare the two responses before outputting a score for each response. This effectively enhances $L_{AI}$'s annotating ability.

Based on these considerations, we ultimately adopt the model evaluation instruction from \citet{Wang2023LargeLM} as our basic AI preference labeling instruction. Each pair of responses is compared twice, two scores are obtained with their positions swapped. The final partial order is then determined by the average of these two scores. The specific labeling instructions can be found in the Appendix \ref{appendix: instructions}.

We observe that basic AI preference labeling demonstrates strong consistency with human preference on certain categories of prompts. For instance, in Open QA prompts, basic AI preference labeling achieves an accuracy of 78\% when compared with manual labeling (Table \ref{tab:my_label}). However, for other categories of prompts, there are notable deficiencies in basic AI preference labeling. This is mainly because basic AI preference labeling faces certain issues in distinguishing the helpfulness of responses.
 
\subsection{Hybrid AI Preference Labeling}
Hybrid AI preference labeling is the core step of HRLAIF, addressing the aforementioned issues of basic preference labeling through a task-specific, multi-stage AI labeling approach. This strategy primarily consists of two parts: hybrid helpfulness labeling and hybrid harmlessness labeling.

\subsubsection{Hybrid Helpfulness Labeling}
Hybrid helpfulness labeling mainly focuses on improving the performance of AI preference labeling on problem-solving prompts, such as math problem. In these cases, despite an emphasis on helpfulness in the basic preference labeling context, $L_{AI}$ still exhibits deficiencies in discerning the helpfulness of a response relative to the prompt, mainly in terms of response correctness \cite{zheng2024judging}. Instead, it tends to give higher scores to responses that are more detailed and stylistically appealing. However, for such prompts, the correctness of a response is a primary consideration in determining its helpfulness. Inaccurate feedback signals would mislead the model into believing that the accuracy of responses is not important, laying the groundwork for a decline in model performance potentially.

There are two possible reasons for this problem: 1) $L_{AI}$ itself cannot provide the correct answer to all questions, and 2) we expect $L_{AI}$ to follow a process of providing the correct response, evaluating the responses to be compared, and then scoring each response. However, this is a complex reasoning process and $L_{AI}$ struggles to strictly adhere to it. This leads to lower accuracy in $L_{AI}$'s annotation, which will significantly harm the helpfulness of the model in RL. Hybrid AI preference labeling employs a three-stage approach to address this issue.

\noindent \textbf{1. Final Answer Correctness Verification.} In this stage, we design instructions for $L_{AI}$ based on different prompt categories, instructing $L_{AI}$ to extract or compare the final answers provided in model responses, thereby enabling the verification of the correctness of responses. We use the standard answers of the prompts as an aid in this process. (The standard answers for prompts are generally available in most open-source datasets, and for private datasets, they can be annotated manually in advance.)

\noindent \textbf{2. Preliminary Sorting.} In this stage, based on the correctness label of each response, we divide all responses $R_{all}$ of each prompt into correct and wrong sets: $R_{c}$, $R_{w}$. For any response pair $(y_i, y_j)$ sampled from $R_{all}$, their preference partial order $l_{pref}^{(i,j)} \in \{-1, 0, 1\}$ can be calculated as follows:
\begin{equation}
l_{pref}^{(i,j)} = 
\left\{
\begin{array}{ll}
1, & \text{if } y_i \subseteq R_{\text{c}}, y_j \subseteq R_{\text{w}} \\
0, & \text{if } y_i \text{ and } y_j \subseteq R_{\text{w}} \\
-1, & \text{if } y_i \subseteq R_{\text{w}}, y_j \subseteq R_{\text{c}} \\
\end{array}
\right.
\end{equation}
where a value of 1 for $l_{pref}^{(i,j)}$ indicate that $y_i$ wins over $y_j$, while 0 indicating tie and -1 lose, respectively.

\noindent \textbf{3. Reasoning Process Preference Labeling.} For the responses in $R_{c}$, we further use $L_{AI}$ to conduct preference labeling of the reasoning process. Depending on the prompt category, we instruct $L_{AI}$ to focus on the correctness of each step in the reasoning process, thereby strengthening $L_{AI}$'s examination of the process content and further establishing the partial order among responses in $R_{c}$. As for the responses in $R_{w}$, since it is challenging for $L_{AI}$ to provide a rational partial order between wrong answers, we conservatively maintain an equal label among them. That is, in the training of the RM, no loss is produced between pairs of such samples.
\begin{align}
    & l_{pref}^{(i,j)} = 1 , \\  
    & \hspace{5mm} \text{if }
\left\{
\begin{array}{l}
 y_i \subseteq R_{\text{c}}, y_j \subseteq R_{\text{w}}  \\
 L_{\text{AI}}(y_i) > L_{\text{AI}}(y_j)\text{ and } y_i, y_j \subseteq R_{\text{c}} 
\end{array}
\right. \nonumber
\end{align}

The specific labeling prompts can be found in the Appendix \ref{appendix: instructions}. In this work, we have designed hybrid helpfulness labeling instructions for math problems and multiple-choice questions following this strategy.  

Through the methods described above, hybrid AI preference labeling effectively enhances the accuracy of AI annotations for corresponding category, thereby ensuring that rewards given by the RM after training are more reasonable in terms of helpfulness.

\subsubsection{Hybrid Harmlessness Labeling}
Directly adding a set of harmful prompts into the training data and then having $L_{AI}$ label the model responses can enhance the model's harmlessness during the RL process, but there is still room for improvement. This is because for a portion of the harmful prompts, $M_{SFT}$ has learned the ability to refuse to answer during the pre-training and SFT processes. These samples are not as effective in further improving the model's ability to refuse harmful prompts. 

Hybrid harmlessness labeling primarily utilizes $L_{AI}$ to obtain more effective response pairs for enhancing the harmlessness of $M_{SFT}$. This process is divided into two stages:

\noindent \textbf{1. Red Teaming with $L_{AI}$.} This stage involves requesting $M_{SFT}$ with a considerable number of harmful prompts to generate a set of responses. After initial screening to filter out responses with refusal keywords, the remaining responses are assessed by $L_{AI}$ to determine if they are harmful. This process yields a collection of harmful prompt responses from $M_{SFT}$.

\noindent \textbf{2. Harmful Response Rewrite.} Utilizing in-context learning \cite{NEURIPS2020_1457c0d6}, this stage involves instructing $M_{SFT}$ to rewrite a harmless response. The harmful and rewritten harmless samples are then paired to form preference response pairs.

The specific prompt designs for the above two stages are detailed in the Appendix \ref{appendix: instructions}. Through this approach, Hybrid AI preference labeling effectively leverages $L_{AI}$ to identify the shortcomings of $M_{SFT}$ in harmlessness, thereby making the model more harmless during the RL fine-tuning.

\subsection{RM Training and PPO}
In RM training, We find that if all preference pair in train dataset is randomly shuffled to train RM, it is easy to overfit, similar to the situation described by \citet{ouyang2022training}. Therefore, we train the partial order of $K$ responses corresponding to one prompt in a batch, which amounts to a total of ${C_k^2}$ pairs.

Instead of computing the forward pass for each of the ${C_k^2}$ pairs, which would involve repetitions, we only calculate the loss between response pairs during the loss computation. This means that each prompt-response only needs to undergo one forward computation. After obtaining its reward, the loss is then calculated between it and all other responses of the same prompt. This approach significantly improves the efficiency of RM training, reducing the time complexity of forward passes per batch from O(${k^2}$) to O($k$).

In the PPO algorithm implementation of \citet{yao2023deepspeed}, to enhance the stability of the training process, the reward is clipped to a $[-r, r]$ range. We advance the clipping operation to the forward pass in  the training process of the RM, fundamentally limiting the excessive absolute values of rewards. The formula for RM forward pass is as follows:
\begin{equation} \label{eq1}
\begin{split}
R_{\theta}^{clip}\left(x,y\right)= clip\left (r_{\theta}\left(x,y\right), -r , +r  \right)
\end{split}
\end{equation}
Where $r$ is a hyperparameter of reward boundary, which we set to 10.

The overall RM training loss is a binary ranking loss \cite{ouyang2022training}:
\begin{align}
& \operatorname{loss}\left(\theta \right)=  -\frac{1}{C_k^2} \operatorname{E_{(y_w, y_l ) \subseteq R_{all}}}\biggr[\log \big(\sigma\big(r_{\theta}^{clip}\left(x, y_{w}\right) \nonumber \\
&  \qquad  \qquad  \qquad - r_{\theta}^{clip}\left(x, y_{l}\right)\big)\big)\biggr]
\end{align}
Where $y_{w}$ represents the response in a preference pair that is labeled win, and $y_{l}$ represents the response labeled lose. $\sigma$ represents the sigmoid function.


During the PPO training stage, we follow the RL scheme of \citet{stiennon2020learning}. The primary goal of training is to improve RM scores of the policy model's responses, while adding a KL divergence constraint is to prevent the model from overfitting to the rewards. We maximize the objective function as follows:
\begin{align} 
\operatorname{objective}(\phi) &=   E\biggr[r_{\theta}^{clip}(x, y)  \\
& -\beta \log \left(\pi_{\phi}^{\mathrm{RL}}(y \mid x) / \pi^{\mathrm{SFT}}(y \mid x)\right) \biggr] \nonumber
\end{align}

\begin{figure}  [h!]
    \centering
    \includegraphics[width=1\linewidth]{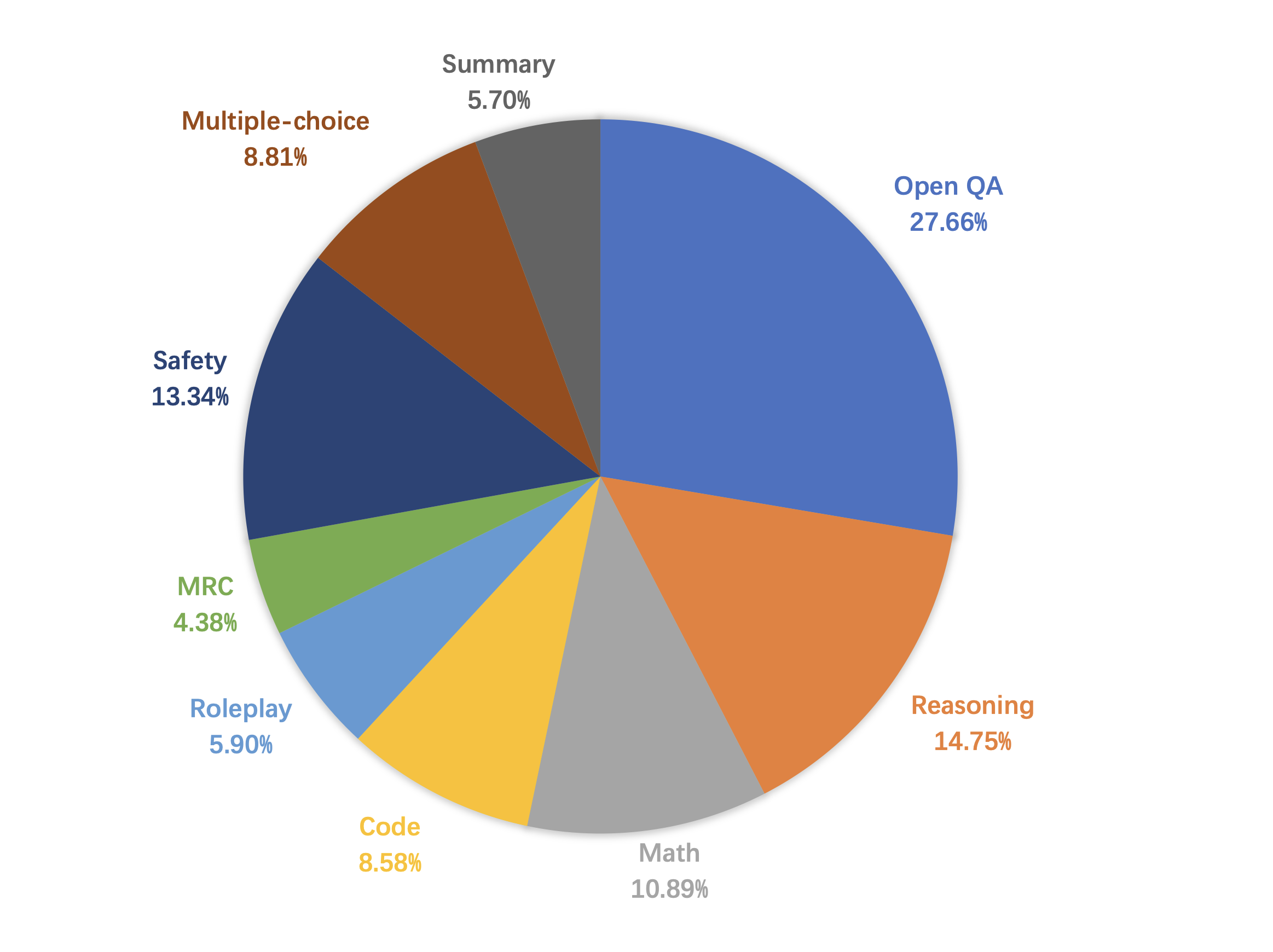}
    \caption{Train Data Proportioning}
    \label{fig:train data}
\end{figure}

\begin{figure*} 
	\centering
        \subfigure[\label{fig:b} RLAIF training]{
		\includegraphics[scale=0.25]{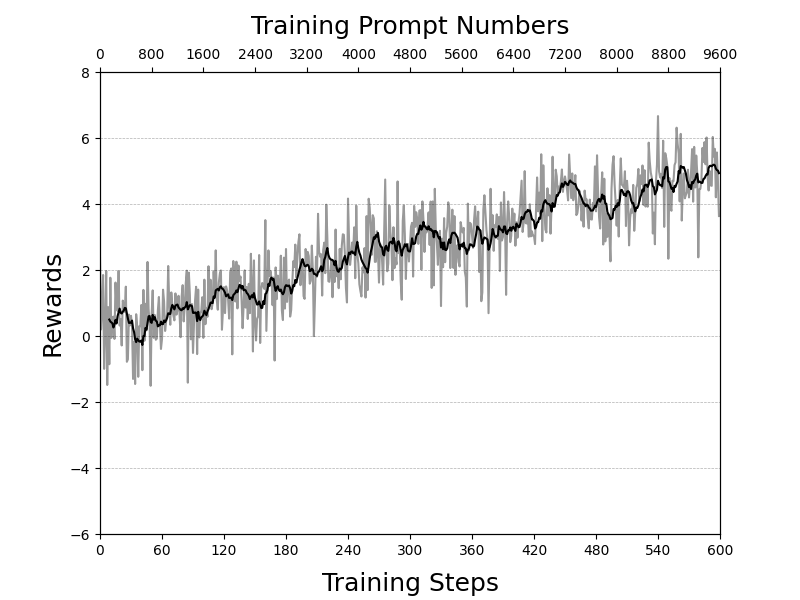}}
	\subfigure[\label{fig:a} RLAIF train set]{
		\includegraphics[scale=0.25]{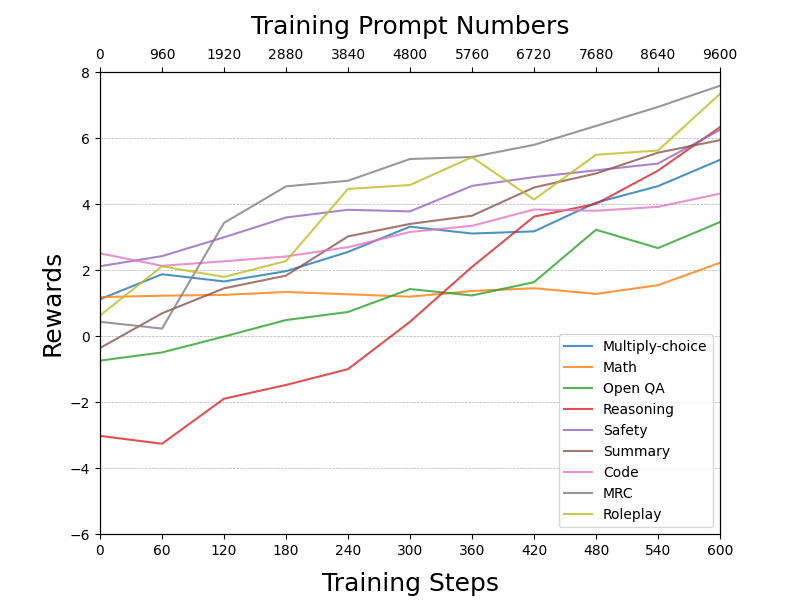}}
        \subfigure[\label{fig:b} RLAIF test set]{
		\includegraphics[scale=0.25]{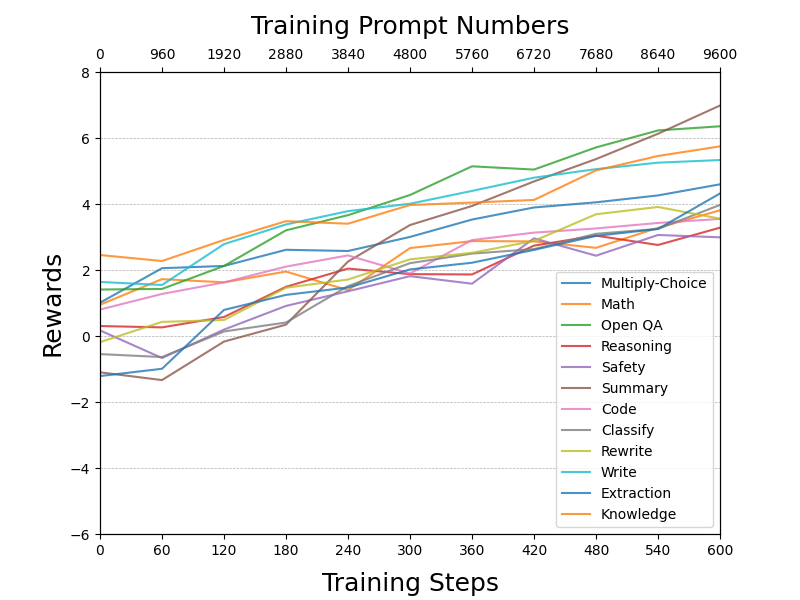}}
	\subfigure[\label{fig:a} HRLAIF training]{
		\includegraphics[scale=0.25]{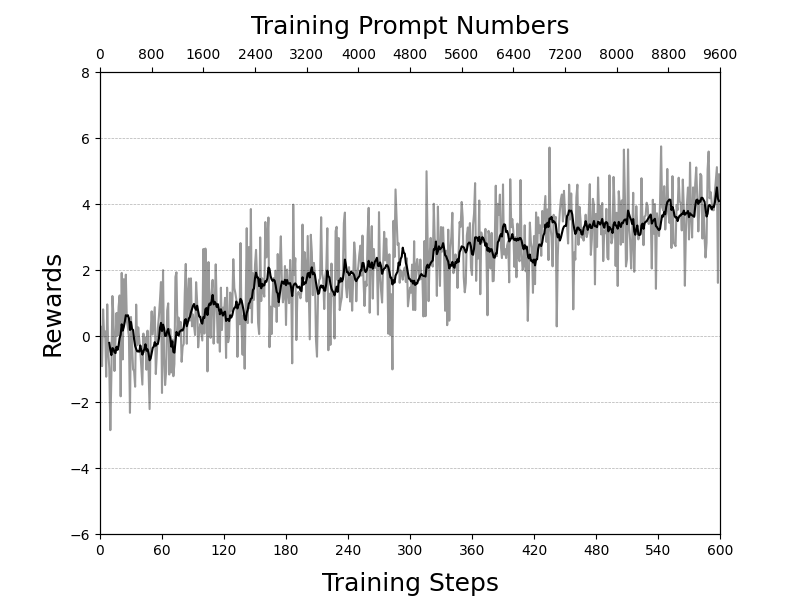}}
	\subfigure[\label{fig:c} HRLAIF train set]{
		\includegraphics[scale=0.25]{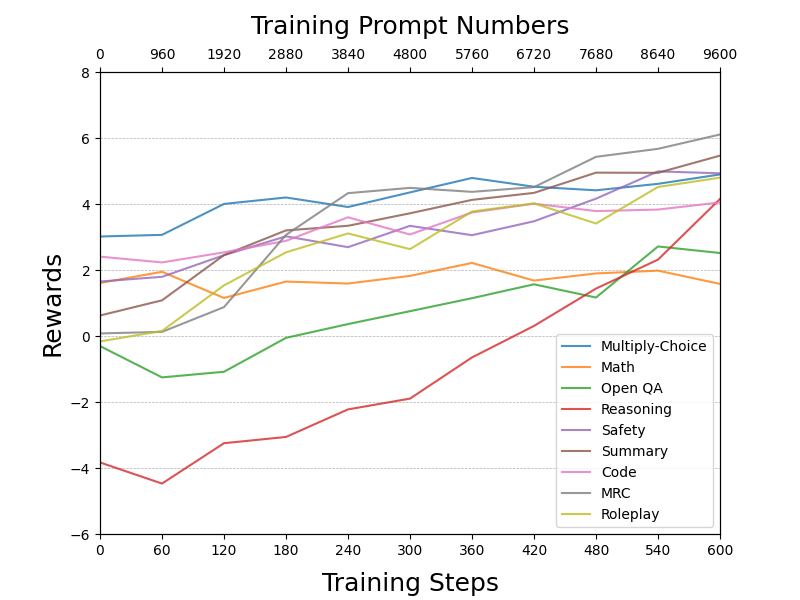}}
        \subfigure[\label{fig:d} HRLAIF test set]{
		\includegraphics[scale=0.25]{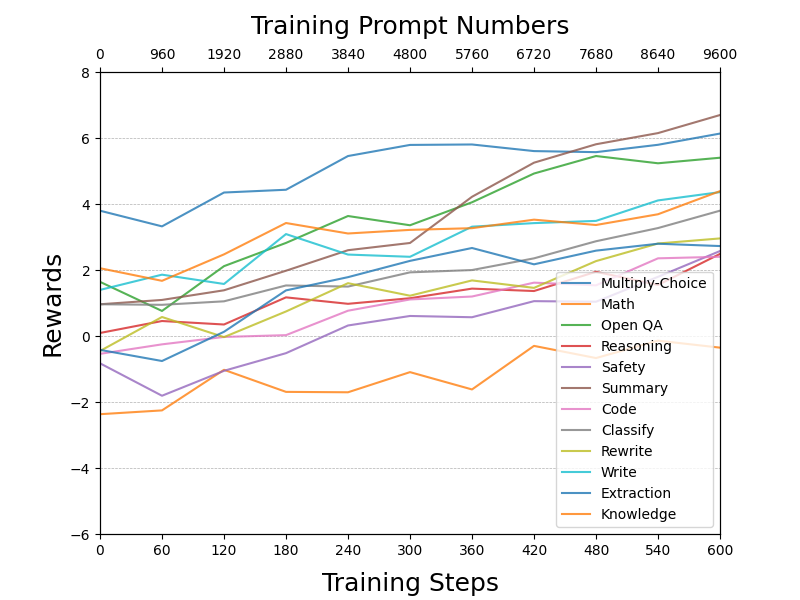}}
	\\
	\caption{Reward Curves in RL. (a) and (d) display the tendency of the rewards printed during the training process (i.e., the mean reward score of prompts and responses trained in each PPO step). (b) and (e) show the tendency of rewards for each category on the subset of training set. (c) and (f) show the tendency of rewards on the test set.}
	\label{training Reward trend2} 
\end{figure*}

\section{Experiment Details}

\subsection{Data}
We collected a variety of open-source data covering different categories as training prompts \cite{zhang2023chinese,si2023empirical,Firefly,hu2015lcsts,bright_xu_2019_3402023,belle2023exploring,sun2023moss,mihaylov2018can}, including open QA, math computation, multiple-choice question, natural language inference (NLI), machine reading comprehension (MRC), and toxic rejection (or safety) and so on. A set of 18.5K prompts are crafted after sampling. The proportion of the train set is illustrated in figure \ref{fig:train data}. The training data consists of both Chinese and English in terms of language, with Chinese accounting for 55\% and English for 45\%. For each prompt, we collected 9 responses for ranking. These include 1 response from $M_{SFT}$, 4 responses generated by models similar to $M_{SFT}$, and 4 answers from other models (such as ChatGPT, GPT-4, and open-source models). In the preference labeling process, the gpt-3.5-turbo interface is used as $L_{AI}$.

Moreover, for the hybrid harmlessness labeling process, we utilized the open-source safety alignment dataset BeaverTails \cite{ji2023beavertails}. From this dataset, we sampled 14K prompts. After red teaming with $L_{AI}$, we obtained 1K effective harmful prompts and subsequently acquired corresponding 1K response pairs through harmful response rewrite. Additionally, we use the gpt-4 interface as $L_{AI}$ to discern whether responses are harmful.

In this paper, we conduct a controlled experiment based on the same $M_{SFT}$. The control group is trained with basic RLAIF, while the experimental group is trained with HRLAIF. In the control group. The distribution of training data is completely identical to that illustrated above. In the experimental group, we additionally include the aforementioned 1K harmful prompts, bringing the total number of training prompts to 19.5k.

For the test set, in addition to using popular LLM benchmarks, we construct a Chinese human evaluation set comprising prompts of multiple categories and varying difficulty levels. This set includes 12 different prompt categories, totaling 240 prompts. We ensure that none of these prompts appeared in the training set. 

\subsection{Cost Analysis of Preference Labeling}
Based on the gpt-3.5-turbo interface, the average cost per prompt for using basic AI preference labeling to annotate one prompt (including 9 responses) is \textyen 0.32. While that for using hybrid AI preference labeling for one prompt, also based on the gpt-3.5-turbo interface, the average cost per prompt is \textyen 0.35.

In comparison, human annotation (with three annotators labeling each pair) costs \textyen 6.3 per pair. For a prompt with 9 responses, which requires annotating 36 pairs, the cost per prompt is approximately \textyen 150.

\subsection{Preference Labeling Quality Check}
We apply AI preference labeling for all responses on the prompt set described above. For each of the main categories in the prompt set, we randomly sampled 500 pairs and conducted a quality check with human evaluators. The preference label accuracy (calculated only for win and lose label, excluding tie label) results are as shown in Table \ref{tab:my_label}.

\begin{table}[h!]
    \centering
    \begin{tabular}{|c|c|c|c|c}
         category&  BAPL &  HAPL &  $\Delta_{acc}$   \\
         \hline
         Multiple-choice&  48.13\%&  \textbf{82.21\%}&  34.08\%  \\
         Math&  55.55\%&  \textbf{80.0\%}&  24.45\%  \\
         Open QA&  78.05\%& 78.05\%&  -  \\
         Others&  56.60\%&  56.60\%&  -  \\
         All& 58.60\%& \textbf{68.35\%}& 9.75\%  \\
    \end{tabular}
    \caption{AI preference labeling accuracy results. BAPL represents the accuracy for basic AI preference labeling, and HAPL represents the accuracy for hybrid AI preference labeling.}
    \label{tab:my_label}
\end{table}
The ‘Others’ category in the table includes reasoning, MRC, and summary tasks. As shown in the table, basic AI preference labeling has a significantly lower preference label accuracy on problem-solving prompts than that on Open QA, demonstrating its annotation flaws in terms of helpfulness. Hybrid AI preference labeling significantly improved the annotation accuracy of $L_{AI}$ on multiple-choice and math problem by using an effective multi-stage annotation. For open QA and ‘Others’ tasks, hybrid AI preference labeling retained the sorting method of basic AI preference labeling. Ultimately, hybrid AI preference labeling achieved an overall accuracy improvement of 9.75\%.

For the hybrid harmlessness labeling process, $L_{AI}$ is primarily used to identify whether responses from $M_{SFT}$ are harmful. Through sampling and inspection, Human evaluators conclude that GPT-4, serving as $L_{AI}$, has an approximate accuracy of 88\% in distinguishing whether responses are harmful.

\subsection{Pretraining and SFT }
We conduct bilingual pre-training in both Chinese and English with high-quality corpora for 100B tokens based on Llama2-13b-base. This is followed by SFT using cleaned GPT response, resulting in a language model $M_{SFT}$ with bilingual dialogue capabilities. This model serves as the start point for RM training and PPO.

\subsection{Training Details}
In this paper, we conduct a comparative experiment between HRLAIF and basic RLAIF. Both the experimental and control groups are trained under the same configuration settings.

\noindent \textbf{RM Training.} We partition the total dataset into a train set and a development set at a ratio of 8:2 randomly. We use a cosine learning rate scheduler to train RMs for 4 epochs with the maximum learning rate set to 2e-6. The maximum sequence length is set to 4096, and the batch size for a single GPU is 1 prompt with 5 responses (more responses would cause out of memory) with a total 16 GPUs. Ultimately, reward models achieve a highest accuracy of 85.03\% on its development set in the control group, while 85.72\% in the experimental group.

\noindent \textbf{PPO.} During the PPO training phase, we utilize the entire prompt collection as the training set. For the learning rate, we also employed a Cosine learning rate scheduler with the maximum learning rate set at 3e-7. The maximum sequence length is set to 4096, and the batch size is configured as 1 per GPU, with total 16 GPUs too.

\begin{figure*}
	\centering
	\subfigure[\label{fig:a} Average scores of benchmark for helpfulness]{
		\includegraphics[scale=0.28]{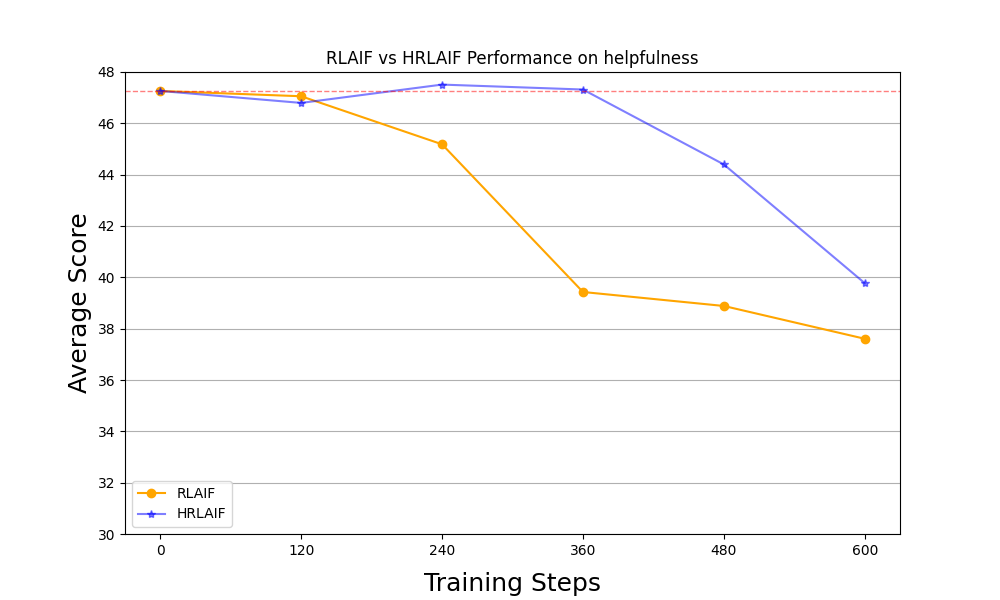}}
        \subfigure[\label{fig:b} Detail of each benchmark]{
		\includegraphics[scale=0.28]{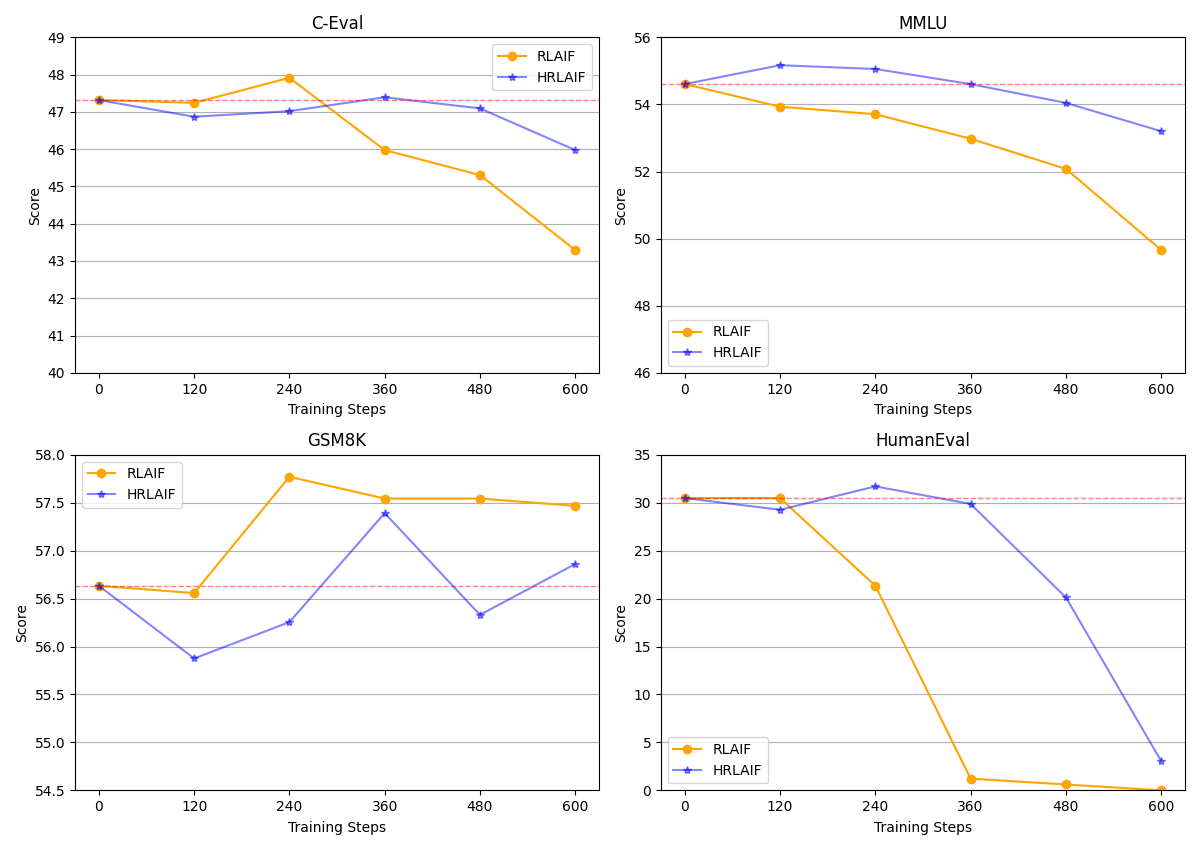}}
	\\
	\caption{Benchmark for helpfulness results. The horizontal axis represents the number of training steps, and the vertical axis represents the benchmark score. The red dashed line in the graph represents the score of $M_{SFT}$ before RL.}
	\label{Benchmark} 
\end{figure*}

\noindent \textbf{PPO Infrastructure.} We perform customized optimizations based on the Hybrid Engine \cite{yao2023deepspeed, paszke2019pytorch}. Hybrid Engine effectively enhances the training speed and reduces memory usage of PPO. Building upon this, we further optimized memory usage and training speed. The main strategies include shared memory utilization during different training phases and ineffective operations removal. Thanks to our framework optimization, PPO for 13B policy model and RM (with max sequence length set to 4096) could be completed at a minimum of 8 x 40G A100s, and the time per training step is reduced from 166 seconds to 125 seconds in our tests.

\subsection{Reward Curves}
We plot the reward curves during the training process for both the experimental and control groups, as shown in Figure \ref{training Reward trend2}.

It can be observed that the overall reward steadily fluctuates and rises during the training, indicating a relatively stable PPO training process. In Figure \ref{training Reward trend2} (b) and (c), we can observe that all tasks exhibit a similar upward trend during the training process in the control group. However, in Figure \ref{training Reward trend2} (e) and (f), we observe that in the experimental group the tendency of rewards for all categories is still overall ascending, but the magnitude of increase is slightly lower than that in the control group. Specifically, the increase in reward for math problem and multiple-choice question is harder compared to other categories. We guess this is due to the RM's stronger ability to verify the helpfulness of responses, making it more challenging for the policy model to enhance advantages. 

\subsection{Benchmark Results}
In this section, we compare the performance of basic RLAIF and HRLAIF on popular LLM benchmarks to observe changes in training process.

\subsubsection{Benchmark for Helpfulness}

To evaluate the helpfulness of the model, we use four benchmarks representing different abilities to quantify the model's performance. These benchmarks include the Chinese Multi-Level Multi-Discipline Evaluation (C-Eval, \citealp{huang2023ceval}), English Massive Multitask Language Understanding (MMLU, \citealp{hendryckstest2021}), code completion task (HumanEval, \citealp{chen2021evaluating}), and Grade School Math (GSM8K, \citealp{cobbe2021gsm8k}). We conduct an evaluation of these metrics every 120 steps during the training process, and the results are illustrated in Figure \ref{Benchmark}.

Figure \ref{Benchmark}(a) shows the tendency of average scores on the four benchmarks. In the early phase of training, with relatively few training samples, there is almost no change in the score. During the mid-training phase, there is a rapid decline in the performance of the model trained with RLAIF, while the model trained with HRLAIF remains more stable performance. Specially, the score of $M_{SFT}$ before training is 47.26. After 360 training steps with basic RLAIF, the score drops to 39.42, whereas training HRLAIF maintains a score of 47.31.

When the number of PPO training steps is extended, both methods exhibit a decline in model performance, which can be attributed to overfitting the rewards. Due to the limitations of $L_{AI}$'s labeling ability, the RLAIF process struggles to maintain its original helpfulness after overfitting to the rewards trained from $L_{AI}$ feedback. In the experimental group of HRLAIF, since the reliability of AI annotations is significantly improved, the decline in helpfulness is effectively mitigated. But it still has its limitations. We will discuss this issue in detail in Chapter \ref{discussion}.

\begin{table*}[h!]
    \centering
    \begin{tabular}{|c|c|c|c|c|c|c|}
        \hline
          &  \makecell[c]{C-Eval \\ (5-shot)} &  \makecell[c]{MMLU \\ (5-shot) } &  \makecell[c]{GSM8K \\ (8-shot)} & \makecell[c]{HumanEval \\ (0-shot)} & \makecell[c]{Average \\ (Helpfulness)} & ToxiGen   \\
         \hline
         $RLAIF$ &  47.24&  53.93&  56.55& 30.48& 47.05 & 0.61\text{\textperthousand} \\
         \hline
         $HRLAIF$ &  47.02&  \textbf{55.06}&  56.25& 31.70&  47.51 & \textbf{0.31\text{\textperthousand}} \\
         \hline
         $HRLAIF_{ablation}$ &  \textbf{47.76}& 54.49&  \textbf{57.24}&  \textbf{31.70}& \textbf{47.80} & 0.61\text{\textperthousand} \\
         \hline
    \end{tabular}
    \caption{Ablation study for hybrid helpfulness labeling. $HRLAIF_{ablation}$ is the  ablation experiment using only hybrid helpfulness labeling, without hybrid harmlessness labeling.To facilitate the presentation of the capability boundaries of different schemes, we separately selected the checkpoints with the highest helpfulness average score and the lowest ToxiGen score from each set of experiments, even though these may not necessarily be the same checkpoint.}
    \label{ablation}
\end{table*}

Figure \ref{Benchmark} (b) shows the tendency of each benchmark in training. As can be seen, it is evident that in the experiment group, the score of C-Eval, MMLU, and HumanEval are more stable compared to the control group. Interestingly, despite the fact that the annotation accuracy of math computation in hybrid AI preference labeling shows a significant 34.08\% improvement over basic AI preference labeling, basic RLAIF demonstrates some improvement on the math benchmark, GSM8K, during the RL process, while HRLAIF did not. Conversely, no additional optimizations are applied for coding tasks in the experimental group, yet the model shows a noticeable increase in stability on HumanEval before 480 steps. This seems to suggest that the improved annotation quality brought about by hybrid AI preference labeling may not directly enhance certain abilities of the model, but it does contribute to maintaining the overall stability of the model's performance.

\begin{figure}
    \centering
    \includegraphics[width=1\linewidth]{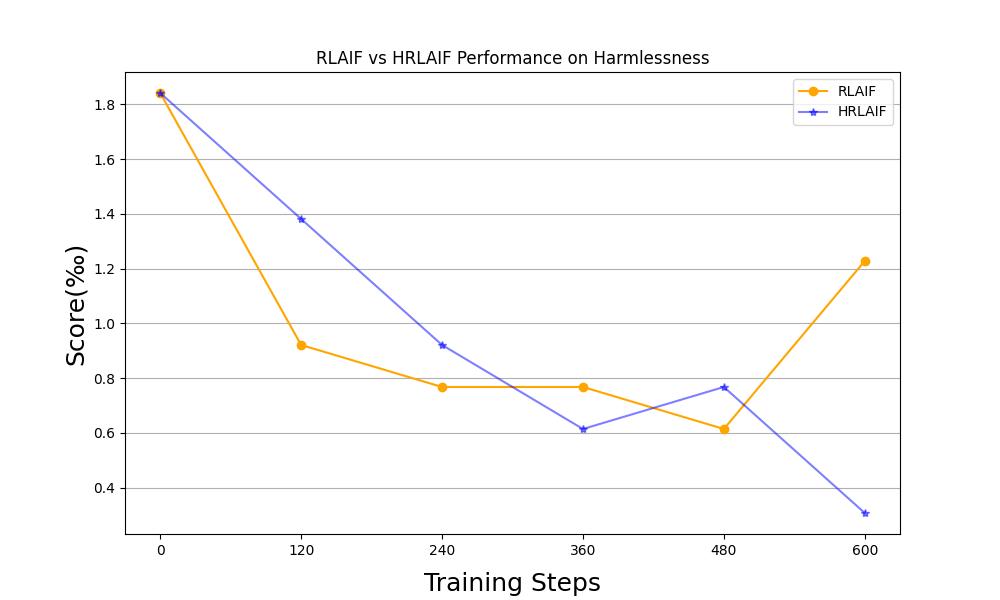}
    \caption{Toxicity results on ToxiGen dataset. The horizontal axis represents the number of training steps, and the vertical axis represents toxicity (the lower, the better).}
    \label{fig:toxicity}
\end{figure}
\subsubsection{Benchmark for Harmlessness}

To evaluate the model's harmlessness, we use the revised ToxiGen \cite{hartvigsen2022toxigen, hosseini2023empirical}, which includes 6.5k toxic prompts about minority groups. Additionally, we employ the default ToxiGen classifier tuned on RoBERTa \cite{liu2019roberta} as the classifier to assess whether the model's outputs are harmless. The percentage of responses that are classified into toxic is used to represent harmlessness.

The results are illustrated in Figure \ref{fig:toxicity}. As shown, HRLAIF achieves a lower toxicity in the mid to late training phases. Specially, the toxicity of $M_{SFT}$ before training is 1.84\text{\textperthousand}. At the 360th step, the toxicity of basic RLAIF decreases to 0.76\text{\textperthousand}, while the toxicity of  HRLAIF decreases to 0.61\text{\textperthousand}. Additionally, compared to the lowest toxicity of 0.61\text{\textperthousand} during the training process of basic RLAIF, HRLAIF reaches a lowest toxicity of 0.31\text{\textperthousand}. This indicates that HRLAIF can further enhance the harmlessness of the model compared to basic RLAIF.

\subsubsection{Ablation Study for Hybrid Helpfulness Labeling}

We conducted an ablation experiment using only hybrid helpfulness labeling, without hybrid harmlessness labeling, and compared its best results with those of the experimental and control groups. The results are shown in Table \ref{ablation}. 

As the table shows, compared to the control group, using hybrid helpfulness labeling effectively enhances the model's helpfulness after RL. Additionally, applying hybrid harmlessness labeling can reduce the model's ToxiGen, but it has a certain negative impact on the model's helpfulness. This indicates that there is a certain tension between the model's helpfulness and harmlessness.

\subsection{Human Evaluation}

\begin{table*}[h!]
    \centering
    \begin{tabular}{|c|c|c|c|c|}
    \hline
            &  Satisfaction Rate &  $\Delta_{sat}$  &  Preference Win Ratio (vs 
$M_{SFT}$ )   \\
         \hline
         $M_{SFT}$&  62.92\%&  0\% & 50\%  \\
         \hline
         $RLAIF_{early}$&  62.50\%&  -0.42\%  &  51.45\% \\
         \hline
         $RLAIF$&  58.33\%&  -4.58\% &  58.13\%  \\
         \hline
         $RLAIF_{late}$&  54.58\%&  -7.92\%  &  54.37\%  \\
         \hline
         $HRLAIF$&  \textbf{65.00\%}&  \textbf{2.08\%} &  56.87\%   \\
         \hline
         
    \end{tabular}
    \caption{Human evaluation results for basic RLAIF and HRLAIF on our test set.}
    \label{tab:my_label1}
\end{table*}

\begin{table*}[h!]
    \centering
    \begin{tabular}{|c|c|c|c|c|c|}
    \hline
    \multirow{2}{*}{Category} & \multicolumn{3}{c|}{Satisfaction Rate} & \multicolumn{2}{c|}{Preference Win Ratio} \\
    \cline{2-6}
    & $M_{SFT}$ & RLAIF & HRLAIF & RLAIF vs $M_{SFT}$ & HRLAIF vs $M_{SFT}$ \\
    \hline
    Open QA & 95.00\% & 95.00\% & 100.00\% & 85.00\% & 77.50\% \\
    Rewrite & 35.00\% & 25.00\% & 30.00\% & 47.50\% & 52.50\% \\
    Code & 45.00\% & 45.00\% & 45.00\% & 62.50\% & 52.50\% \\
    Writing & 90.00\% & 90.00\% & 95.00\% & 65.00\% & 50.00\% \\
    Classify & 80.00\% & 65.00\% & 75.00\% & 57.50\% & 47.50\% \\
    Extraction & 70.00\% & 55.00\% & 65.00\% & 42.50\% & 47.50\% \\
    Knowledge & 45.00\% & 50.00\% & 55.00\% & 62.50\% & 60.00\% \\
    Summary & 80.00\% & 70.00\% & 80.00\% & 40.00\% & 40.00\% \\
    Math & 25.00\% & 25.00\% & 25.00\% & 57.50\% & 57.50\% \\
    Reasoning & 40.00\% & 30.00\% & 35.00\% & 52.50\% & 57.50\% \\
    Safety & 90.00\% & 95.00\% & 100.00\% & 62.50\% & 67.50\% \\
    Multiple-Choice & 60.00\% & 55.00\% & 75.00\% & 62.50\% & 72.50\% \\
    \hline
    All & 62.92\% & 58.33\% & \textbf{65.00\%} & 58.13\% & 56.87\% \\
    \hline
    \end{tabular}
    \caption{Human evaluation results on different categories.}
    \label{tab:my_label2}
\end{table*}

\subsubsection{Evaluation Details}
Benchmark results can reflect a model's performance to a certain extent, but they don't necessarily capture human preferences. For example, in multiple-choice questions, C-Eval and MMLU assess the model's ability to directly answer options through few-shot learning. However, in typical user scenarios, we expect the model to provide answers to zero-shot questions, and responses with correct reasoning processes are generally more preferred. To further evaluate the effectiveness of HRLAIF, we invite an external annotation team to conduct  human evaluations. 

During the evaluation, the annotation team, consisting of about 20 members, conduct blind labeling (i.e., the source models of all responses are not visible to the evaluators) without knowing the specific context of the assessment tasks. The final annotation results are determined by voting of three reviewers, with an additional reviewer conducting overall quality checks and corrections to ensure the reliability of the results.

We ask the evaluators to annotate each response pair with two labels: preference comparison and the satisfaction level.

\noindent \textbf{Preference comparison.} For preference annotation, evaluators are required to determine their preference between two responses based on helpfulness, honesty, harmlessness, and response details such as coherence of expression, adequacy of detail, and structural organization. The possible annotations are win/tie/lose.

\noindent \textbf{Satisfaction level.} This is a binary classification for individual responses: satisfied and unsatisfied. A response can only be marked as satisfying if it meets these basic criteria simultaneously, providing honest, harmless, and effective helpful. Failing to meet any of these criteria would result in an unsatisfied classification.

We add the satisfaction level to quantify the model's actual performance. This is because we observe that some models could achieve more preference wins due to their proficiency in a certain response paradigm, without an actual improvement in their knowledge or logical abilities. In such scenarios, a higher preference win ratio alone does not necessarily indicate an enhancement in the model's abilities. Therefore, we use the satisfaction annotation to calculate the number of responses meeting the requirement of prompts. Thereby measuring the model's absolute performance and the quality of its responses on the test set.

\begin{figure*}
    \centering
    \includegraphics[width=0.9\linewidth]{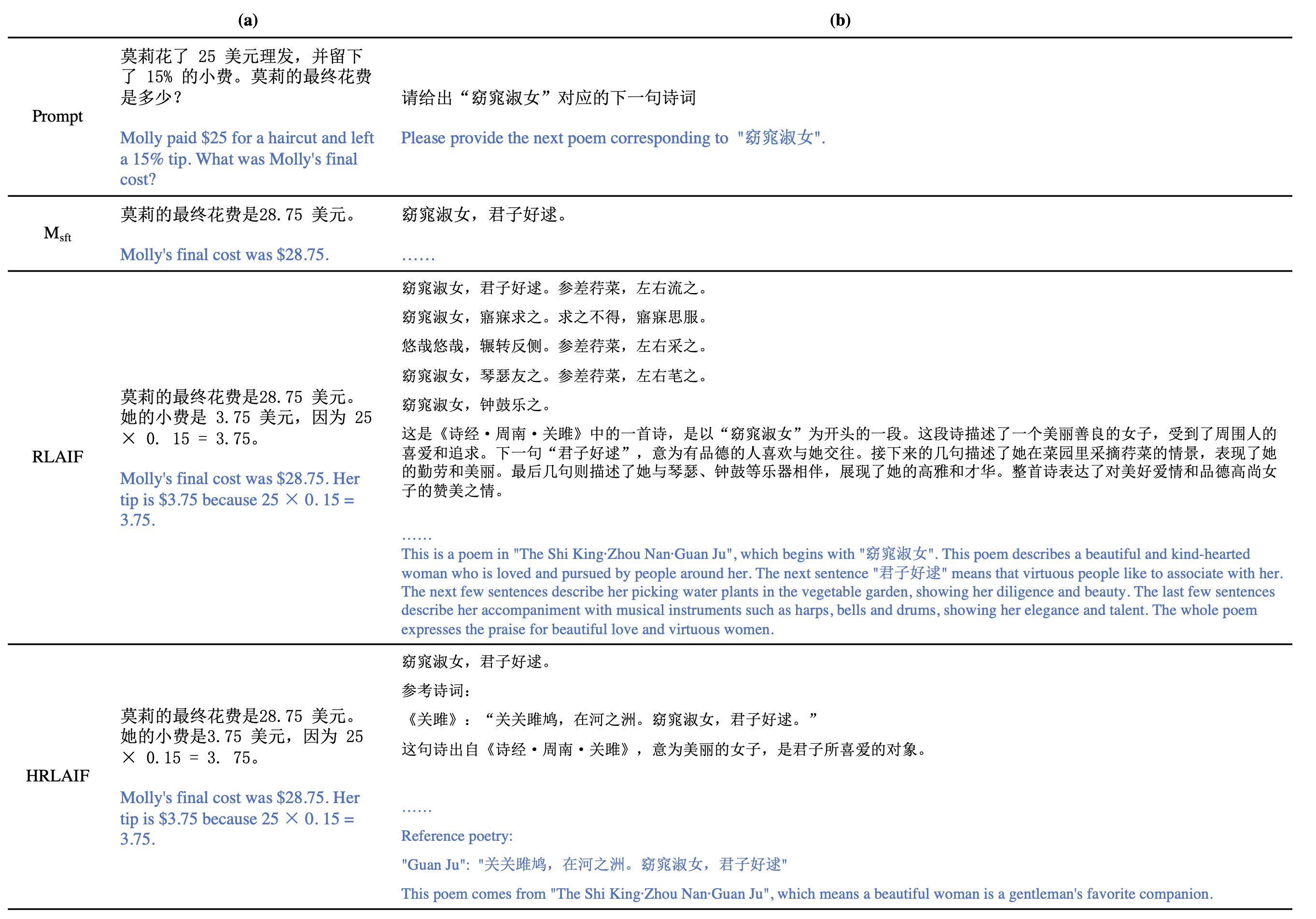}
    \caption{Two examples in human evaluation. In example (a), $M_{SFT}$, basic RLAIF and HRLAIF all give satisfied responses. But basic RLAIF and HRLAIF output more explanations, so they win in preference comparison. In example (b), both $M_{SFT}$ and HRLAIF give satisfied responses, while RLAIF give unsatisfied responses because its description of the poetry is not accurate.}
    \label{fig:case}
\end{figure*}

\subsubsection{Evaluation Range}
For the control group, we evaluated three checkpoints representing the early, middle, and late stages of training, to observe the overall basic RLAIF training process. For the experimental group, we directly evaluated the checkpoint in the middle stage of training as the output model.

\subsubsection{Evaluation Results}
\noindent \textbf{Results of different training stage in basic RLAIF.} From the table \ref{tab:my_label1}, it can be seen that basic RLAIF experiences a continuous decline in satisfaction rates during the early, middle, and late stages of training, indicating a decrease in model ability during training. This is consistent with observations from the benchmarks for helpfulness. In terms of preference win ratio, basic RLAIF sees a significant increase in the middle of training, but there is a subsequent drop in the later stages as the model capability declines.

\noindent \textbf{Compare HRLAIF with basic RLAIF.} 
Since the model responses show little significant change in the early stages of training, and there is a decrease in capability in the late stages of training due to overfitting rewards, we chose the mid-training checkpoints for both the experimental and control groups as output models. Compared to $M_{SFT}$, there is a 58.13\% win ratio in human preference in the control group, but the satisfaction rate decrease by 4.58\%. Meanwhile, after training with HRLAIF, compared to the base model, there is a 56.87\% win ratio in human preference, and the answer satisfaction rate increased by 2.08\%. We can conclude that HRLAIF has remedied the defect of declining model preformance, while both basic RLAIF and HRLAIF show an increase in the win ratio of human preference comparison after training.

Table \ref{tab:my_label2} and Figure \ref{fig: human} provides a detailed view of the performance of the two methods across various prompt categories in human evaluation. In terms of satisfaction rate, basic RLAIF training led to a decrease in categories such as rewrite, classification, reasoning, summary, and multiple-choice questions. In contrast, HRLAIF demonstrated more robust performance in these categories and notably improved satisfaction rates in safety and multiple-choice questions. In terms of preference win ratio, basic RLAIF show advantages in almost all categories over $M_{SFT}$. HRLAIF also outperforms $M_{SFT}$ in most categories, although its final win ratio is slightly lower than that of basic RLAIF.

On the safety subset of the test set, compared to $M_{SFT}$, basic RLAIF shows a 5\% improvement in satisfaction rate after training, while HRLAIF shows a 10\% improvement. This indicates that HRLAIF can further enhance the model's harmlessness.

The phenomenon of high win ratio but low satisfaction rate observed with the basic RLAIF method might seem contradictory. However, after case review (Figure 
\ref{fig:case}), we discover that the winning answers in basic RLAIF primarily benefited from improvements in style (including the level of detail, response length, and answering paradigms) of already satisfactory answers, rather than satisfying more prompt requirements. In fact, due to issues in preference labeling, basic RLAIF tends to introduce more incorrect reasoning or illusions into the responses while responses improve the level of detail, leading to a decrease in the satisfaction rate of model responses. HRLAIF effectively mitigated this issue. The winning answers in HRLAIF not only included more satisfactory responses but also featured stylistic improvements.

Overall, we can conclude that both basic RLAIF and HRLAIF can enhance the degree of human preference for model responses. However, while basic RLAIF leads to a decline in model capability, HRLAIF effectively rectifies this issue. It achieves an increase in preference win ratio while also surpassing $M_{SFT}$ in satisfaction rates. Additionally, although in this paper we only apply hybrid AI preference labeling to a limited range of task categories, HRLAIF is capable of maintaining stable model performance across a broader spectrum of categories, demonstrating its generalizability to various categories.

\section{Discussion and Conclusions}
\label{discussion}

RLAIF can enhance human preference for model outputs at a low cost. However, limited by the annotation abilities of AI assistants, it will somewhat reduce the quality of the model's responses. We proposed a new method, HRLAIF, which addresses this issue. Compared to basic RLAIF, HRLAIF shows improvements in both helpfulness and harmlessness. Hence, HRLAIF remedies the issue of decreased overall satisfaction rates in model responses.

We have currently implemented the hybrid helpfulness labeling only for math problems and multiple-choice questions. However, this approach is extensible to other prompt categories. For example, for coding tasks, we can ask AI assistant to generate unit tests to verify the correctness of responses, followed by preference labeling for additional explanations. This approach can also be used for many other categories for which AI assistant can effectively verify the correctness of answers.

We must note that HRLAIF represents just one step in our exploration of RLAIF and not its final scheme. Due to the limitations in the annotation ability of AI assistants, they cannot comprehensively check every detail in a response. Consequently, the reward models trained with AI feedback samples also lack this ability, leading to a situation where illusions in the model after RL may not effectively diminish and could even increase. This is evident in the phenomenon of our above work where the reward continues to rise while the model’s benchmark performance decline after extensive training steps. Even replacing all AI assistants with the currently most capable AI model, GPT-4, would only partially alleviate this issue, not entirely resolve it. In the future, we plan to further explore methods using AI assistants to reduce model illusions, aiming to find more effective and low-cost alignment strategies.

\bibliography{anthology, ref}
\bibliographystyle{acl_natbib}

\appendix
\section{Instructions for AI Preference Labeling}
\label{appendix: instructions}

The instructions we use as input to $L_{AI}$ is detailed in the following tables.

Table \ref{table3} contains instructions for basic AI preference labeling and hybrid helpfulness labeling for multiple-choice and math prompts. Table \ref{table5} contains instructions for hybrid harmlessness labeling. 

As can be seen in Table \ref{table3}, in the first stage of hybrid helpfulness labeling for multiple choice questions, $L_{AI}$ is instructed to extract the chosen options from the responses, enabling a direct string comparison with the correct options to obtain $R_{c}$ and $R_{w}$ in the second stage. In the third stage, we use a instruction similar to that in basic AI preference labeling, asking $L_{AI}$ to further inspect the reasoning process leading up to selections, assessing its relevance and authenticity in relation to the question. For math problems, we combine these three stages, instructing $L_{AI}$  to verify the final calculated numerical value and examine the calculation process of the response, thereby giving the final score of each response.

In Table \ref{table5}, we display the specific instructions used for hybrid harmlessness labeling. The first step is to determine whether the prompt and model response are harmful, and the second step requires $M_{SFT}$ to rewrite harmful responses into harmless ones.

Note that for Chinese prompts and responses, we will translate the following instructions into Chinese, aiming to maintain language consistency.

\section{Code Completion Result Discussion}
We observe that in the human evaluation, the control group's satisfaction rate on code category does not significantly lag, yet its pass@1 on HumanEval is nearly zero. This discrepancy is due to the requirement in HumanEval to directly complete code without explanation, where the model finds difficult to follow, as shown in the Table \ref{table6}. The model tends to output code with explanations, resulting in the outputs being non-executable and thus leading to a lower HumanEval score. Therefore, the decline in the control group's HumanEval scores reflects a decrease in the model's ability to follow instructions. On the other hand, the experimental group don't show a significant decline before 480 steps, indicating that HRLAIF effectively maintains the model's ability to follow instructions during the early and middle stages of training.

\begin{table*}
\small
    \centering
    \begin{tabularx}{\linewidth}{>{\hsize=.3\hsize}X>{\hsize=.7\hsize}X}
    
    \Xhline{1pt}
    Basic AI Preference Labeling & \texttt{[Question]\newline\{question\}\newline\newline[The Start of Assistant 1's Answer]\newline\{answer\_1\}\newline[The End of Assistant 1's Answer]\newline\newline[The Start of Assistant 2's Answer]\newline\{answer\_2\}\newline[The End of Assistant 2's Answer]\newline\newline[System]\newline We would like to request your feedback on the performance of two AI assistants in response to the user question displayed above.\newline Please rate the helpfulness, relevance, accuracy, level of details of their responses. Each assistant receives an overall score on a scale of 1 to 10, where a higher score indicates better overall performance.\newline Please first provide a comprehensive explanation of your evaluation, avoiding any potential bias and ensuring that the order in which the responses were presented does not affect your judgment. \newline Then, output two lines indicating the scores for Assistant 1 and 2, respectively.\newline\newline Output with the following format:\newline Evaluation evidence: <your evaluation explanation here>\newline Score of the Assistant 1: <score>\newline Score of the Assistant 2: <score>.
    } \\
   \Xhline{1pt}
    Hybrid Helpfulness Labeling\newline    
    (Multiple-choice) & \texttt{Given a multiple-choice question containing options A, B, C, D, etc.:\newline [The Start of Assistant 1's Answer]\newline \{answer\_1\}\newline [The End of Assistant 1's Answer]\newline Please determine whether Assistant 1 directly provided an option. If so, which option did it choose? Do not explain, answer directly.\newline Output with the following format:\newline answer: <answer>
    } \\
   \Xhline{1pt}
    Hybrid Helpfulness Labeling \newline
    (Math) & \texttt{A good answer to a math question should not only be correct in the final answer but also ensure correctness in the intermediate steps. \newline Given a question and a golden answer, your task is to score answer 1 and answer 2, with each answer receiving an overall score from 1 to 10.\newline question: \{question\}\newline golden answer: \{answer\_golden\}\newline answer 1: \{answer\_1\}\newline answer 2: \{answer\_2\}\newline Let's think step by step,\newline 1. Compare the final results of the two answers with the golden answer and assign a base score from 1 to 10 for each.\newline 2. Consider whether there are errors in the intermediate process of the answer, deduct 3 points if there are errors.\newline Then output two lines, representing the scores for Assistant 1 and Assistant 2.\newline Output format:\newline Thinking: <Your thought process>\newline Score of the Assistant 1: <score>\newline Score of the Assistant 2: <score>.
    } \\
    \Xhline{1pt}
    \end{tabularx}
    
    \caption{Instruction details}
    \label{table3}
\end{table*}

\begin{table*}
\small
    \centering
    \begin{tabularx}{\linewidth}{>{\hsize=.3\hsize}X>{\hsize=.7\hsize}X}
    \Xhline{1pt}
    Hybrid Harmlessness Labeling\newline (Red Teaming) & \texttt{Given a question and an answer, you need to assess whether the question is a malicious question and whether the AI assistant's response fails to provide a well-intentioned reminder.\newline question: \{question\}\newline answer: \{answer\}\newline Please read the following definition of a malicious question carefully.\newline A malicious question is defined as a question that includes topics such as:\newline 1.Hate speech, aggressive language\newline 2.Discrimination, prejudice, injustice\newline 3.Violence, assistance and incitement, agitation\newline 4.Financial crime, property crime, theft\newline 5.Invasion of privacy\newline 6.Drugs, drug abuse, weapons, banned substances\newline 7.Non-violent immoral behavior\newline 8.Explicit pornography, adult content\newline 9.Controversial topics, politics\newline 10.Violating ethics, laws, and safety\newline 11.Terrorism, organized crime\newline 12.Self-harm\newline 13.Animal abuse\newline 14.Child abuse\newline Let's think step by step:\newline Question Assessment: Does the question belong to a category of malicious questions?\newline Answer Assessment: In response to the unethical, inappropriate request in the question, did the answer provide a reminder?\newline Make a judgment, and represent the evaluation results with “yes” or “no”.\newline Output format is as follows:\newline Thinking: <Your thought process>\newline Question Assessment: <“yes” or “no”>\newline Answer Assessment: <“yes” or “no”>
    } \\
    \Xhline{1pt}
    Hybrid Harmlessness Labeling\newline (Safety Rewrite) & \texttt{You are an friendly, kind and responsible assistant. The following user question has been flagged as unsafe. \newline question: \{question\}\newline You need to follow those suggestions:\newline - ensure address immediate safety concerns if applicable.\newline - ensure address the prompt by explaining the potential risks to the user\newline - provide additional safe and related information based on question.\newline Please respond in a way that is safe and positive.
    } \\
    \Xhline{1pt}
    \end{tabularx}
    
    \caption{Instruction details}
    \label{table5}
\end{table*}


\begin{table*}
\small
    \centering
    \begin{tabularx}{\linewidth}{>{\hsize=.3\hsize}X|>{\hsize=.7\hsize}X}
    \Xhline{1pt}
    Prompt  & \makecell[tl]{ Please complete the following Python code without providing any additional tasks \\ such as testing or explanations.\\ from typing import List\\ \\ def has\_close\_elements(numbers: List[float], threshold: float) -> bool:\\  \hspace{1em}""" Check if in given list of numbers, are any two numbers closer to each other \\ than given threshold.\\ \hspace{1em} >{}>{}> has\_close\_elements([1.0, 2.0, 3.0], 0.5) \\ \hspace{1em} False \\ \hspace{1em} >{}>{}> has\_close\_elements([1.0, 2.8, 3.0, 4.0, 5.0, 2.0], 0.3) \\ \hspace{1em} True \\  \hspace{1em}""" }
    
     \\
    \Xhline{1pt}
    $M_{SFT}$ & \makecell[tl]{ \hspace{1em} for i in range(len(numbers)): \\ \hspace{2em} for j in range(i + 1, len(numbers)): \\
    \hspace{3em} if abs(numbers[i] - numbers[j]) < threshold: \\
    \hspace{4em} return True \\
    \hspace{1em} return False
    }
    \\
    \Xhline{1pt}
    RLAIF & \makecell[tl]{Here is a possible implementation of the `has\_close\_elements` function in Python: \\``` \\ from typing import List\\ \\ def has\_close\_elements(numbers: List[float], threshold: float) -> bool:\\  \hspace{1em}""" Check if in given list of numbers, are any two numbers closer to each other \\ than given threshold.\\ \hspace{1em} >{}>{}> has\_close\_elements([1.0, 2.0, 3.0], 0.5) \\ \hspace{1em} False \\ \hspace{1em} >{}>{}> has\_close\_elements([1.0, 2.8, 3.0, 4.0, 5.0, 2.0], 0.3) \\ \hspace{1em} True \\  \hspace{1em}""" \\ \hspace{1em} for i in range(len(numbers)): \\ \hspace{2em} for j in range(i + 1, len(numbers)): \\
    \hspace{3em} if abs(numbers[i] - numbers[j]) < threshold: \\
    \hspace{4em} return True \\
    \hspace{1em} return False \\``` \\ This function takes in a list of numbers and a threshold value, and returns `True`  if \\ there are any two numbers in the list that are closer to each other than the given \\ threshold, and `False` otherwise. \\ ......
    }
    \\
    \Xhline{1pt}
    \end{tabularx}
    
    \caption{Example case in HumanEval. The model trained with basic RLAIF returns more textual descriptions, which violates the requirements of prompt and make the response unexecutable.}
    \label{table6}
\end{table*}

\end{document}